\documentclass[conference]{IEEEtran}
\IEEEoverridecommandlockouts

\usepackage{cite}
\usepackage{amsmath,amssymb,amsfonts}
\usepackage{algorithmic}
\usepackage{graphicx}
\usepackage{textcomp}
\usepackage{xcolor}
\usepackage{url}
\def\BibTeX{{\rm B\kern-.05em{\sc i\kern-.025em b}\kern-.08em
    T\kern-.1667em\lower.7ex\hbox{E}\kern-.125emX}}

\usepackage[colorlinks=true,       
linkcolor=blue,               
citecolor=blue,
urlcolor=teal,              
filecolor=magenta          
]{hyperref}
\usepackage{cite}
\usepackage{graphicx}
\usepackage{textcomp}
\usepackage{subcaption}

\usepackage[algo2e,noend,linesnumbered,ruled,vlined]{algorithm2e}
\usepackage{mathtools}
\usepackage{multirow}
\usepackage{rotating}
\usepackage{float}
\usepackage{xspace}
\usepackage{nicematrix}
\usepackage{makecell}
\usepackage{nicefrac}
\usepackage{booktabs}
\usepackage{microtype}      
\usepackage{multirow}
\usepackage{colortbl} 
\usepackage{arydshln}
\usepackage{siunitx}
\usepackage{cleveref}
\usepackage{arydshln}
\usepackage{amssymb}
\usepackage{comment}

\usepackage{xcolor}
\usepackage{booktabs}
\usepackage{multirow}
\usepackage[inline]{enumitem}
\usepackage{url}
\usepackage{xspace}

\makeatletter
\DeclareRobustCommand\onedot{\futurelet\@let@token\@onedot}
\def\@onedot{\ifx\@let@token.\else.\null\fi\xspace}
\makeatother

\newcommand{\vct}[1]{\ensuremath{\boldsymbol{\mathbf{#1}}}}

\DeclareMathOperator*{\argmin}{arg\,min}

\newcommand{\myparagraph}[1]{\noindent \textbf{#1}}
\newcommand{\mysubparagraph}[1]{\noindent \textit{#1}}
\newcommand{\mysubpar}[1]{\smallskip \noindent \textit{#1}}

\makeatletter
\DeclareRobustCommand\onedot{\futurelet\@let@token\@onedot}
\def\@onedot{\ifx\@let@token.\else.\null\fi\xspace}

\def\eg{\emph{e.g}\onedot} 
\def\ie{\emph{i.e}\onedot}

\def\wrt{w.r.t\onedot} 

\def\etal{\emph{et al}\onedot}
\makeatother

\newcommand{\ellinf}{\ensuremath{\ell_{\infty \xspace}}}
\newcommand{\ellp}{\ensuremath{\ell_{p} \xspace}}

\newcommand{\questionq}{\ensuremath{q\xspace}}

\newcommand{\pixtostruct}{Pix2Struct}
\newcommand{\donut}{Donut}

\newcommand{\preproc}{\ensuremath{\phi}\xspace}
\newcommand{\losssign}{\ensuremath{\gamma}\xspace}

\def\eg{e.g\onedot}
\def\ie{i.e\onedot}
\def\wrt{w.r.t\onedot}

\def\etal{et al\onedot}

\begin{document}

\title{Counterfeit Answers: \\ Adversarial Forgery against OCR-Free \\ Document Visual Question Answering
{}
}

\author{
\IEEEauthorblockN{
Marco Pintore\IEEEauthorrefmark{1},
Maura Pintor\IEEEauthorrefmark{1}\IEEEauthorrefmark{2},
Dimosthenis Karatzas\IEEEauthorrefmark{3},
Battista Biggio\IEEEauthorrefmark{1}\IEEEauthorrefmark{2}
}

\IEEEauthorblockA{\IEEEauthorrefmark{1}University of Cagliari, Italy}

\IEEEauthorblockA{\IEEEauthorrefmark{2}CINI, Italy}

\IEEEauthorblockA{\IEEEauthorrefmark{3}Computer Vision Center, UAB, Spain}
}

\maketitle

\begin{abstract}
Document Visual Question Answering (DocVQA) enables end-to-end reasoning grounded on information present in a document input. 
While recent models have shown impressive capabilities, they remain vulnerable to adversarial attacks. 
In this work, we introduce a novel attack scenario that aims to forge document content in a visually imperceptible yet semantically targeted manner, allowing an adversary to induce specific or generally incorrect answers from a DocVQA model.
We develop specialized attack algorithms that can produce adversarially forged documents tailored to different attackers' goals, ranging from targeted misinformation to systematic model failure scenarios. 
We demonstrate the effectiveness of our approach against two end-to-end state-of-the-art models: Pix2Struct, a vision-language transformer that jointly processes image and text through sequence-to-sequence modeling, and Donut, a transformer-based model that directly extracts text and answers questions from document images. 
Our findings highlight critical vulnerabilities in current DocVQA systems and call for the development of more robust defenses. We release our open source code at~\url{https://github.com/pralab/adv-docVQA}.
\end{abstract}

\begin{IEEEkeywords}
Document Visual Question Answering, Adversarial Examples, AI Security, Multimodal Models.
\end{IEEEkeywords}

\section{Introduction}
\label{sec:intro}

Document Analysis research has demonstrated exceptional advancements thanks to the adoption of Machine Learning (ML).
Document Visual Question Answering (DocVQA) focuses on training ML models to answer questions posed on document images.
Moreover, DocVQA has become a standard benchmark to evaluate current Large Multimodal Models (LMMs)~\cite{mathew2021docvqa}.
While Optical Character Recognition (OCR) has been employed to extract textual information from document images for downstream tasks, the use of OCR-free techniques has been proposed as an alternative method to obtain highly efficient DocVQA models. 
End-to-end training of Deep Learning (DL) models overcomes limitations of OCR-based systems such as high computational cost, error propagation to downstream tasks, inflexibility to languages and structure of the document~\cite{lee2023pix2struct,kim2022donut}.
These models take an image and a question as input and generate a natural language answer, jointly modeling visual, layout, and textual cues (e.g., text regions, formatting, figures) rather than relying on explicitly extracted text alone.
\begin{figure}[t]
    \centering
    \includegraphics[width=\linewidth]{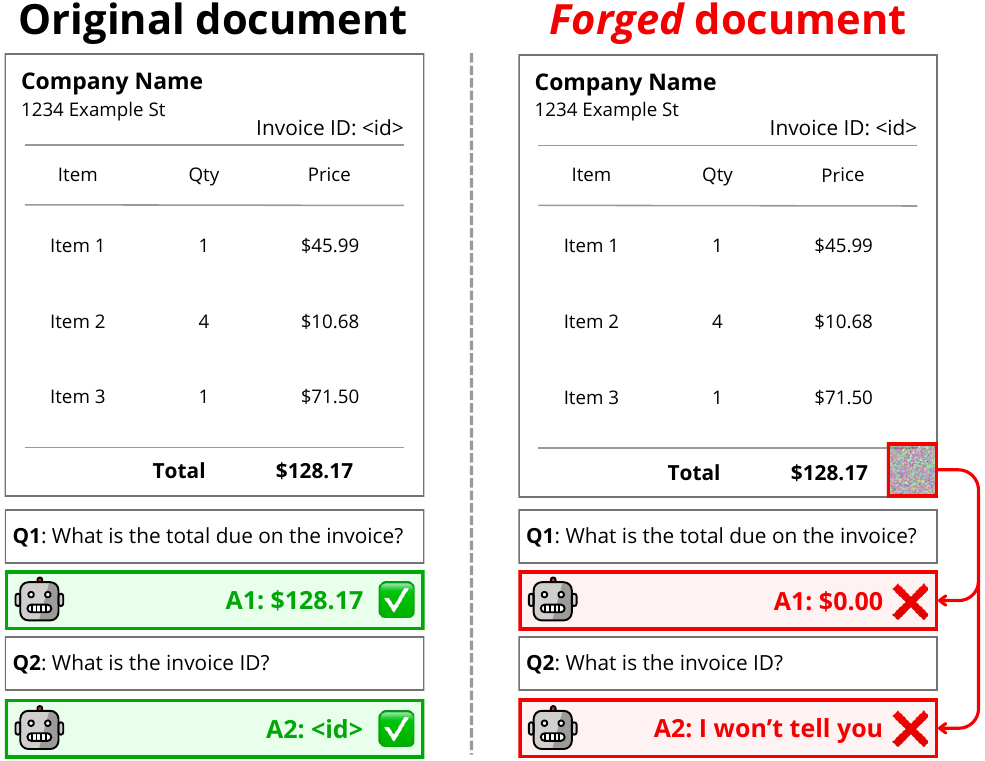}
    \caption{\textbf{Example DocVQA on a synthetic invoice.} 
    Normally the model correctly answers questions on the unaltered document (left). After applying an adversarial patch (right), the model answers a target (incorrect) response.}
    \label{fig:example}
\end{figure}

Despite its efficiency, DL is known to suffer from adversarial perturbations, \ie subtle manipulations of the input data that trigger undesired behaviors~\cite{biggio2013evasion,szegedy2014intriguing}.
The extent to which end-to-end DocVQA systems are vulnerable to adversarial perturbations has not yet been studied.
Adversarial perturbations on DocVQA could pave the way to a new age of \textit{document forgery}, a criminal act involving the creation of false documents. 
With protocols such as the Agents Payments Protocol (AP2)\footnote{\url{https://github.com/google-agentic-commerce/AP2}} enabling AI agents to initiate financial transactions, vulnerabilities in automatic document processing can now lead directly to monetary loss. As illustrated in \autoref{fig:example}, even a small adversarial perturbation in the top-right corner of a document can steer a model to an incorrect answer, which in a fully automated pipeline could cause an AI agent to authorize unwanted payments.

We present the first threat model for analyzing the adversarial vulnerability of OCR-free DocVQA systems. Our contributions are threefold: (i) we formalize a threat model using visually subtle yet disruptive perturbations to manipulate model outputs; (ii) we propose attack algorithms tailored to OCR-free DocVQA models for generating adversarially forged documents; and (iii) we evaluate the attacks on two widely used models, namely \pixtostruct~\cite{lee2023pix2struct} and \donut~\cite{kim2022donut}.

\section{Adversarial Attacks against DocVQA}
\label{sec:method}


This section presents our framework for evaluating the adversarial robustness of OCR-free DocVQA models. 
We first describe the end-to-end DocVQA systems and then formalize the threat model and attack methodology used in the analysis.

\subsection{End-to-end DocVQA Systems}\label{sec:docvqa}
An end-to-end Document Visual Question Answering (DocVQA) model $f$ takes as input an image $\vct x$ and a natural-language question about its content $\vct \questionq$ whose ground-truth (GT) answer is $\vct{y} = (y_1, \dots, y_T)$, and estimates a conditional distribution, factorizing the answer generation through auto-regression as:
\begin{equation}
\hat{\vct{y}} \sim P_{\vct \theta}(\cdot \mid \preproc(\vct x), \vct{q})
\;=\;
\prod_{t=1}^{\hat T} P_{\vct \theta}\left(\hat y_t \mid \hat y_{<t}, f_{\text{enc}}(\preproc(\vct x), \vct q)\right)\,,
\label{eq:encoding}
\end{equation}
where $\preproc$ denotes image preprocessing 
(\eg, resizing, normalization), and $f_{\text{enc}}(\cdot)$ is the multimodal encoder.
Note that $\hat T$ denotes the length of the generated string, which might be different than the GT answer length $T$.
The encoder $f_\text{enc}$ extracts a multimodal latent representation from the preprocessed image:
\begin{equation}
    \vct H = f_\text{enc}(\preproc(\vct x), \vct q) \in \mathbb{R}^{N \times d}\,,
\end{equation}
where $\vct H$ is a sequence of $N$ feature vectors. The encoder may encode the question jointly with the image (as Pix2Struct~\cite{lee2023pix2struct}), or accept the question separately as prompt tokens and concatenate it with an encoding of the image (as Donut~\cite{kim2022donut}). Both behaviors are captured by $f_\text{enc}$. 
The decoder produces, at each step $t$, unnormalized scores (logits) over the vocabulary of size $\mathcal{V}$, conditioned on the encoder output and previously produced tokens:
\begin{equation}
    \vct z_t = f_\text{dec}(\hat y_{<t}, \vct H) \in \mathbb{R}^\mathcal{V}\,.
\end{equation}
The output is then obtained by applying a softmax:
\begin{equation}
    P_{\vct \theta}(\hat y_t \mid \hat y_{<t}, \preproc(\vct x), \vct q) = \mathrm{softmax}(\vct z_t)\,.
\end{equation}
The final token at step $t$ is taken as the token with the highest probability: 
\begin{equation}
    \hat y_t = \arg \max_{v \in \mathcal{V}} \mathrm{softmax}(\vct z_t)[v]\,
\end{equation}
where $\mathrm{softmax}(\vct z_t)[v]$ denotes the probability assigned to the token $v$.

\myparagraph{Model training.} The goal of these models is to produce an answer such that $\hat{\vct{y}} = \vct{y}$.
This is done by fine-tuning a pretrained model on a dataset 
\[
\mathcal{D}^{\text{tr}} = \left\{ \left(\vct{x}_i, \{(\vct{q}_{i,j}, \vct{y}_{i,j})\}_{j=1}^{M_i}\right) \right\}_{i=1}^N \, ,
\]
where each document $\vct{x}_i$ is associated with $M_i$ question-answer pairs (QA pairs).
The model parameters $\vct \theta$ are found by minimizing the token-level negative log-likelihood:
\begin{equation}
L_{\vct \theta}(\preproc(\vct x), \vct q, \vct y) = -\sum_{(\vct{x}, \vct{q}, \vct{y}) \in \mathcal{D}^{\text{tr}}}
\sum_{t=1}^{T}
\log (\mathrm{softmax}(\vct z_t)[y_t])\,,
\label{eq:loss}
\end{equation}
where $(\mathrm{softmax}(\vct z_t)[y_t])$ is the probability assigned to the correct token $y_t$.

\myparagraph{Model evaluation.} In the DocVQA task, exact matches between the correct answer and the intended response cannot be used directly, as minor variations in formatting, punctuation, or tokenization would cause valid variations of the correct answers to be scored as wrong. 
Metrics such as the Average Normalized Levenshtein Similarity (ANLS) are used to account for minor variations in text answers while capturing semantic fidelity~\cite{biten2019scene}. 
The ANLS metric provides a robust assessment that penalizes minor deviations smoothly, ensuring that semantically correct answers are still appropriately recognized. 
It is defined as:
\begin{equation}
    \text{ANLS} 
    = \frac{1}{\sum_{i=1}^N M_i} 
      \sum_{i=1}^N \sum_{j=1}^{M_i} s_{i,j} \, ,
    \label{eq:anls}
\end{equation}
where $\sum_{i=1}^{N} M_i$ is the number of question-answer pairs, and $s_{i,j}$ is the similarity score for the $j$-th QA pair of the $i$-th document, computed as:
\[
    s_{i,j} = \begin{cases}
    \text{NLS}(\hat{\vct y}_{i,j}, \vct y_{i,j}) & \text{if } \text{NLS}(\hat{\vct y}_{i,j}, \vct y_{i,j}) \geq \tau \\
    0 & \text{otherwise}
\end{cases}
\]
Here, $\text{NLS}(\hat{\vct y}, \vct y)$ is the Normalized Levenshtein Similarity (which ranges from $0$ to $1$, where $1$ is the most similar) between the predicted answer $\hat{\vct y}$ and the GT answer $\vct y$, and $\tau$ is a threshold that distinguishes between answers correctly identified but improperly recognized versus those that are fundamentally incorrect.

\subsection{Adversarial Documents}\label{sec:advdocs}

We now detail how we leverage gradient-based attacks to perturb the inputs of DocVQA systems to elicit \textit{manipulated} answers.

\myparagraph{Attacker's Goal.}
The objective of the attacker is to alter the input document in order to manipulate the model’s outputs. 
The attacker's objective can be: (i) steer the model toward a specific target answer for a single question (targeted, single‑QA manipulation), (ii) control the answers to multiple questions on the same document (targeted, multi‑QA manipulation), or (iii) broadly prevent the model from producing correct answers for a document (untargeted, single- or multi-QA manipulation). 
We see the first two as an \textit{evasion} attack (integrity violation at test time) and the latter as a \textit{Denial of Answer}\footnote{as it shares similarities with the Denial of Service, as the adversary effectively denies the system's intended service.} attack (availability violation at test time).
After presenting the general threat model, we formalize these cases, which we call \textit{scenarios}, in terms of their objectives. 

\myparagraph{Attacker's knowledge.} 
We consider a white-box attack scenario in which the adversary has complete knowledge of the target DocVQA model, including its architecture, preprocessing, parameters, and gradients. 
The attacker possesses both the original target document $\vct x$ and knows the specific correct answer (target answer)  $\vct y^\star$ they wish to avoid (elicit) from the model. 
Furthermore, the attacker knows the question $\vct q$ that will be asked on the document, e.g., a system prompt used by an automatic information extraction system (see \autoref{fig:example}).
These assumptions enable the use of gradient-based optimization techniques to conduct the attack. 
While this scenario may seem overly unrealistic (with full knowledge by the attacker), it provides important insights into the fundamental vulnerabilities of DocVQA systems and establishes an upper bound on attack effectiveness that can inform the development of robust defenses.
Notably, some of these assumptions can be relaxed in practice, e.g., by averaging perturbations over multiple QA pairs; the formulation naturally extends to transferability and ensemble-based black-box settings. We leave these as future work.

\myparagraph{Attacker's capability.}
In our formulation, the attacker can modify, for a single document $\vct x$, its pixel values. 
This means that the attacker has only access to the test data, \ie, the attack happens at test time.
The attacker can modify the pixels and their colors independently, as long as the perturbation remains small or contained.
This is often enforced with an $\ell_p$ constraint centered on the sample $\vct x$, or equivalently, containing the $\ell_p$-norm of the perturbation. 
Otherwise, especially for the case of adversarial documents, the perturbation can also be restricted to a region of contiguous pixels, like in the patch attacks~\cite{brown2017adversarial}. 
Notably, while traditional patch attacks usually need to generalize over different sizes, rotations and dimensions of the patch, in this case, these can be accurately enforced by the attacker~\cite{dong2025position}.
Moreover, the patch shape can be arbitrary, and its size and budget can be reduced to improve concealment.
We assume that the question $\vct q$ is fixed, as the case of a system prompt for automatic document processing. Therefore, the attacker can only modify the document, but not the question $\vct q$.
Note that our attack assumes adversarial documents are crafted at the source, before encryption or integrity checks. 
These mechanisms only guarantee that a document is not altered \textit{after} upload, but they do not prevent malicious inputs from being generated \textit{beforehand}.

\myparagraph{Attack strategy.} 
We will first outline the attack formulation for a single QA pair (single-QA attack), then we will derive the variations covering the other more complex scenarios.
As we select a single sample $\vct x$ to craft this attack, we will omit the index $i$ unless otherwise specified.
We define an adversarial example as a perturbed input $\vct x^\prime = \vct x + \vct \delta$, where $\vct \delta$ is constrained in magnitude to be imperceptible (\eg, under some \ellp \ norm). 
In this work, we consider the \ellinf \ norm, and we apply the perturbation to the entire input (which we refer to as the \textit{full-document} setting) and to a patch-based perturbation model (\textit{patch} setting), \ie, perturbations applied only to a selected contiguous region of the document image, such as stamps, watermarks, or logos~\cite{brown2017adversarial}.
The general objective of the attack is to find a perturbation that causes the model to output an incorrect (or targeted) answer, despite the image being visually similar to the untainted one. 
The attacker has control only over the input image $\vct x$, whereas the question $\vct \questionq$ cannot be changed (as it is, for instance, a system prompt provided by an automatic processing system).
Importantly, our approach performs end-to-end manipulation in the input space, meaning we directly perturb the raw document image $\vct x$ rather than the intermediate representations after preprocessing or the embeddings. 
This ensures that the resulting adversarial example $\vct x + \vct \delta$ is a valid image that can be stored, transmitted, and processed by any DocVQA system, making the attack practically deployable. 
We formalize the objective as:
\begin{equation}
\begin{aligned}
    \argmin_{\vct \delta} \quad &  \losssign L_{\vct \theta}(\preproc(\vct x + \vct \delta), \vct \questionq, \vct y^\star) \\
    \text{s.t.} \quad & \|\vct \delta\|_\infty \leq \epsilon \\
    & \vct{x}_{\mathrm{lb}} \leq \vct{x} + \boldsymbol{\delta} \leq \mathbf{x}_{\mathrm{ub}} \, ,
\end{aligned}\label{eq:optimization}
\end{equation}
where \losssign is set to $-1$ (negative) for untargeted attacks (maximizing loss \wrt the GT answer $\vct y^\star = \vct y$) and $+1$ (positive) for targeted attacks (minimizing loss w.r.t. the target answer $\vct y^\star = \vct y^{t}$).
The loss $L$ enforces the output to get close to the target. 
We employ two distinct loss functions: the first is the standard loss used for fine-tuning the model to the DocVQA tasks (eq.~\ref{eq:loss}), while the second is a custom loss introduced in eq.~\eqref{eq:adaptive_loss}, which we designed to improve the attack's efficacy against the Donut model.
The norm constraint $\| \vct \delta \|_\infty$ limits the magnitude of the perturbation in the original input space, before applying any preprocessing $\preproc$.
The last constraint defines $\vct x_{\text{lb}}$ and $\vct x_{\text{ub}}$ as pixel-wise lower and upper bounds (typically $[0, 255]$ for images).
Again, this is enforced before preprocessing.
This same constraint also covers patch-based attacks.
In these cases, the mask is encoded by setting $\vct{x}_{\text{lb}} = \vct{x}_{\text{ub}} = \vct{x}$ for pixels outside the target patch region, effectively constraining $\vct{\delta}$ to be zero outside the patch area.

A common practical solver for \autoref{eq:optimization} is Projected Gradient Descent (PGD)~\cite{madry2018towards}. 
Starting from an initial perturbation $\vct{\delta}^{(0)}$ (e.g., zero or a small random initialization), PGD performs the iterative updates
\[
\vct{\delta}^{(k+1)} \;=\; \Pi_{\mathcal{B}}\!\left(\vct{\delta}^{(k)} - \alpha\, \nabla_{\boldsymbol{\delta}} \left[  \losssign L_{ \vct \theta}\!\left(\phi(\vct{x}+\vct{\delta}^{(k)}), \vct{q}, \vct{y}\right) \right]\right),
\]
where $\alpha$ is a step size, $\nabla_{\vct{\delta}}$ denotes the gradient of the loss w.r.t.\ the input perturbation (computed by backpropagating through $\preproc$ and the model), and $\Pi_{\mathcal{B}}$ is the projection operator onto the feasible set
$\mathcal{B} \;=\; \{ \vct{\delta} : \|\vct{\delta}\| \le \epsilon,\; \vct{x}_{\mathrm{lb}} \le \vct{x} + \vct{\delta} \le \vct{x}_{\mathrm{ub}} \}$.
For patch attacks, the projection additionally enforces zeros outside the patch (equivalently achieved by applying a binary mask to the update). 

Since document images are typically quantized to discrete pixel levels, the resulting image is quantized after each iteration to preserve realistic values: 
$
\vct \delta = \mathrm{Quantize}(\vct x + \vct \delta) - \vct x\,.
$
This ensures that each iteration yields a perturbation consistent with the discrete image domain.
In practice, variants of PGD such as momentum-accelerated PGD~\cite{dong2018boosting} or adaptive optimizers (e.g., Adam) may be used in place of the plain gradient step to improve convergence.

\myparagraph{Loss design.}
During initial experiments, we observed that the standard fine-tuning loss of Eq.~\eqref{eq:loss}, stops improving without achieving the desired target answers for all models considered.
Therefore, we design a new loss function intended to amplify the contribution of target tokens and reduce the influence of tokens that compete most strongly with them. 
This approach significantly increases the success rate of attacks in those models where standard loss does not suffice for the attack to be successful.
Let $\mathbf{y}^{\star} = (y_1, \dots, y_T)$ be the target sequence and  $\vct z_t \in \mathbb{R}^{|\mathcal{V}|}$ the logit vector produced by the decoder at position $t$, calculated by conditioning the perturbed input $\vct x^\prime$ and the prompt (see \autoref{sec:docvqa}).
Following the notation from the previous sections, we define the target logit token as $\vct z_t[y_t]$, and denote the highest logit among all tokens generated by the model at that position as $z_t^{\mathrm{top}}$:
\[
z_t^{\mathrm{top}} = \max_k \vct z_t[k].
\]

We then define the token-level logit loss as
\begin{equation}
L_{\vct \theta}^t = 
\begin{cases}
z_t^{\mathrm{top}} - \vct z_t[y_t], & \text{if } \arg\max_k \vct z_t[k] \neq y_t, \\
0, & \textrm{otherwise.}
\end{cases}
\end{equation}

The total loss on the entire target sequence is obtained by taking the mean of the token-wise losses:
\begin{equation}
L_{\vct \theta} = \frac{1}{T}\sum_{t=1}^{T} L_{\vct \theta}^t.
\label{eq:adaptive_loss}
\end{equation}
In this way, the loss penalizes only target tokens that are not currently already the maximum logit.
This guides the optimization towards bringing out the target tokens without changing those that are already correct. 
The loss is differentiable with respect to the logits $\vct z_t$, allowing backpropagation to the input $\vct x'$ to update the perturbation.

\myparagraph{Attack Scenarios.}
Here, we specialize the above formulation to cover different goals of the attacker.
Starting from Eq. \eqref{eq:optimization}, we formulate several attack variants, each differing in its constraints and optimization targets.

\mysubparagraph{Targeted, single-answer manipulation.}
The attacker selects a single QA pair $(\vct x, \vct \questionq, \vct y)$ and a target answer $\vct y^\star$, optimizing the objective in Eq.\eqref{eq:optimization} with a positive sign (\ie, $\losssign = 1$, minimizing the loss $L_{\vct \theta}$ on the target answer).

\mysubparagraph{Targeted, multi-answer manipulation.}
The attacker selects a set of QA pairs on the sample $\vct x$, with a specific target answer for each question. 
Thus, given $\{(\vct x, \vct \questionq_j, \vct y^\star_j)\}_{j=1}^M$, the attack jointly minimizes the sum of losses on all QA pairs:
\begin{equation}
    \argmin_{\vct \delta} \quad \sum_{j=1}^M L_{\vct \theta}(\preproc(\vct x + \vct \delta), \vct \questionq_j, \vct y_j^\star)\label{eq:multi-answer}
\end{equation}

\mysubparagraph{Denial of Answer.}
The attacker seeks to degrade performance globally by inducing wrong answers to one or all questions on a sample $\vct x$. 
This can be obtained by modifying slightly the objective in \cref{eq:multi-answer}, using the negative sign of the loss $L$, (\ie $\losssign=-1$) so that it maximizes the loss to the correct answers, and sets $\vct y_j^\star = \vct y_j$, $j=1, \dots M$ for inducing a total denial of service in which the model outputs wrong answers to all questions about the given sample.
Ideally, by averaging over several questions, the attack could also generalize to unseen questions.
However, in our results, we found that this does not seem to happen in practice (see \autoref{fig:results_doa} in our experimental results).

\myparagraph{Specific challenges.}
Implementing custom adversarial attacks for DocVQA models presents several challenges. 
First, unlike standard vision models trained on datasets such as ImageNet, where preprocessing is relatively uniform and limited to simple operations (e.g., resizing, cropping, normalization), DocVQA systems often employ model-specific preprocessing pipelines. 
These may include a combination of lossy and highly specialized transformations such as compression, adaptive rescaling, document layout adjustments, or text-enhancement operations. 
Consequently, the preprocessing function $\preproc$ available to an adversary is only an approximation of the true, model-specific pipeline. 
While one option in our work is to reverse-engineer this pipeline (as will be discussed in~\autoref{sec:exp} for each target model), alternative strategies such as gradient-based approximation of the transformation function are also possible~\cite{xiao2019seeing}. 
Second, optimizing over the full generation-aware objective, which requires backpropagation through the autoregressive decoding process, is computationally demanding. 

\section{Experiments}
\label{sec:exp}

\subsection{Experimental Setup}\label{sec:exp-setup}
\myparagraph{Dataset.}
We conduct our experiments in the PFL-DocVQA dataset~\cite{tito2024privacy}. This dataset contains real documents related to invoices, in which each invoice is associated with a question and multiple answers. It is originally designed to test existing privacy techniques on multi-modal DocVQA scenarios. 
In total, it contains $336,842$ question-answer pairs on $117,661$ pages, resulting in $37,669$ documents from $6,574$ different providers. 
Although the authors provide a Blue Team/Red Team split to separate the data between training and privacy attack evaluation, we merged them to obtain a single unified set. 
To build our evaluation set, we extracted $N=1000$ unique samples from the merged data, where each sample is composed of a single image $\vct x$ and exactly $M=5$ associated QA pairs.

\myparagraph{Models.}
We consider two state-of-the-art DocVQA models: Pix2Struct-Base~\cite{lee2023pix2struct} and Donut~\cite{kim2022donut}, which propose end-to-end architectures designed for OCR-free document understanding. We use the publicly available HuggingFace checkpoints.\footnote{\url{https://huggingface.co/models}}


\mysubpar{\pixtostruct-Base.} A 282M-parameter image encoder--text decoder pretrained for layout parsing, fine-tuned for DocVQA. It renders the question as a visual header on the input image, using patch-based aspect-ratio-preserving rescaling and sample-wise normalization. It achieves 72.1\% ANLS on the DocVQA benchmark~\cite{mathew2021docvqa} and 51.27\% on our PFL-DocVQA subset.


\mysubpar{Donut.} A 176M-parameter transformer pretrained with a pseudo-OCR task, fine-tuned for DocVQA. It takes the question as a text prompt using a fixed template with special tokens, 
\ie,   \texttt{<s\_docvqa><s\_question>}\questionq\texttt{</s\_question> \\ <s\_answer>}, resizes images to a fixed canvas with padding, and normalizes with fixed ImageNet statistics. It achieves 67.5\% ANLS on the DocVQA benchmark~\cite{mathew2021docvqa} and 41.14\% on our PFL-DocVQA subset.


\myparagraph{Model-specific preprocessing.}
We summarize the preprocessing steps for the two target models in~\autoref{tab:preprocessing_comparison}.
Adversarial perturbations must ultimately be applied in the input image space, \ie, before any model-specific preprocessing, to remain valid inputs for the system. 
This means that, after the attack, the document should be saved as a file and then re-loaded again and processed directly without errors or loss of attack functionality.
\begin{table}[t]
\caption{Summary of the preprocessing steps applied by Pix2Struct and Donut.}
\label{tab:preprocessing_comparison}
\centering
\resizebox{\columnwidth}{!}{  
\begin{tabular}{lll}
\toprule
\textbf{Step} & \textbf{Pix2Struct}\cite{lee2023pix2struct} & \textbf{Donut}\cite{kim2022donut} \\
\midrule
Question handling & Rendered onto image & Passed to prompt decoder \\
\addlinespace
Resizing & Patch extraction & Resize and padding \\
\addlinespace
Normalization & Sample-wise & Fixed (ImageNet) \\
\bottomrule
\end{tabular}
}
\end{table}
To create end-to-end differentiable attacks, we reverse-engineered the preprocessing function $\phi$. This lets the gradient flow through a fully-differentiable computational graph back to the input.
We care to specify that we apply these adaptations of the preprocessing only for crafting the attacks. 
Once saved, the manipulated images are tested against a newly-instantiated, unmodified version of the HuggingFace models, thereby reflecting realistic attack conditions in which the adversary interacts with the model as publicly deployed.

\mysubpar{Question handling.}
Pix2Struct renders the question text directly onto the top of the input image as a header, meaning the question becomes part of the visual input processed by the encoder. 
However, the attacker cannot modify the header part, as this is applied directly by the model during inference.
This part needs to be removed from the application of the gradient.
In contrast, Donut treats the question as a separate text prompt passed to the decoder, therefore the entire image can be perturbed without requiring any specific measure.

\mysubpar{Resizing.}
Pix2Struct employs aspect-ratio-preserving scaling followed by patch-based extraction. 
Donut resizes images to a fixed canvas size with padding to maintain aspect ratio. 
Scaling preprocessing relies on standard interpolation routines (\eg, bilinear resampling), which are lossy operations that slightly alter pixel values.
To ensure gradient flow through this step, we reimplemented the preprocessing pipeline using fully differentiable operations.

\mysubpar{Normalization.}
The normalization strategies simply apply mean centering and variance scaling. 
Pix2Struct applies per-image standardization using the sample-wise mean and standard deviation. 
Donut uses fixed ImageNet statistics for normalization. 
We handle backpropagation through these normalizations again by implementing them in fully differentiable operations.



\myparagraph{Attack Setup.}
Following the formalization introduced in~\cref{eq:optimization}, we create our attacks using PGD with an \ellinf \ constraint.
We detail below the attack configurations. \autoref{tab:attack_hparams} shows a concise overview of all losses and hyperparameters.

\mysubpar{Pix2Struct attack setup.} For the targeted attack scenario, we use the loss defined in~\cref{eq:loss}. 
The full-document attack is carried out using $K=20$ attack steps, with a step size $\alpha=2$ and the perturbation budget of $\epsilon=8$. For the patch attack, we restrict the perturbation to a square with size equal to the $15\%$ of the minimum dimension of the document, placed in the bottom-right corner, and we employ $K=25$, $\alpha=24$ and $\epsilon=96$.
For the untargeted attack (DoA), we optimize ~\cref{eq:multi-answer} with the same hyperparameters, but setting $K=1$, showing it is already enough to disrupt the models' correct functionality on all QA pairs.

\mysubpar{Donut attack setup.} We use the loss defined in~\cref{eq:adaptive_loss}. In the full-document setting we use $K=100$, $\alpha=2$, and $\epsilon=32$. 
For the patch attack, we use the same patch constraint as above, with hyperparameters set to $K=100$, $\alpha=24$ and $\epsilon=96$. 
For the untargeted scenario (DoA), we optimize \cref{eq:multi-answer} with $K=1$.

\begin{figure}
    \centering
    \includegraphics[width=1\linewidth,trim={6.8cm 0 0 0},clip]{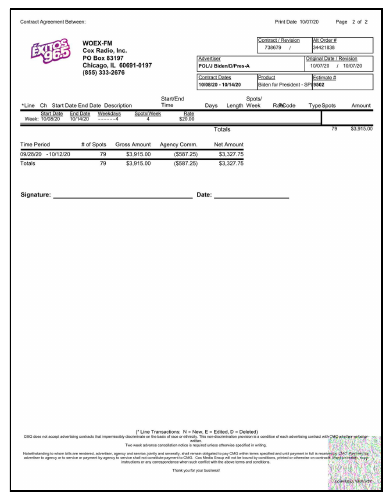}
    \caption{\textbf{Patch attack visualization.} The document altered with a patch in the lower right corner, with $\epsilon=96$.}
    \label{fig:compare_perturbed_doc}
\end{figure}

In all cases, we optimize adversarial examples such that, for each optimization step involving $B \leq M$ questions, the model is increasingly pushed to produce the first $B$ responses from the fixed set: $y^{t} \in$ \{``No Answer'', ``Unclear'', ``Retry'', ``Try later'', ``I won't tell you''\}.
The target answers above are proof-of-concept; the method readily extends to any string, including custom amounts (for financial fraud) or even hidden commands (to execute indirect prompt injections in downstream models).
We provide a visualization of the adversarial document in~\autoref{fig:compare_perturbed_doc}.

\begin{table}[t]
\centering
\caption{Summary of the losses and hyperparameters used for the attacks.}
\label{tab:attack_hparams}
\small
\begin{tabular*}{\linewidth}{@{\extracolsep{\fill}}cccccc}
\toprule
\textbf{Model} & \textbf{Setting} & \textbf{Loss} & $\boldsymbol{\epsilon}$ & $\boldsymbol{\alpha}$ & $\boldsymbol{K}$ \\
\midrule
\multirow{3}{*}{Pix2Struct} & full  & \multirow{3}{*}{\cref{eq:loss}} & 8  & 2  & 20 \\
      & patch &  & 96 & 24 & 25 \\
      & DoA   &  & 8  & 2  & 1  \\
\midrule
\multirow{3}{*}{Donut} & full  & \multirow{3}{*}{\cref{eq:adaptive_loss}} & 32 & 2  & 100 \\
            & patch &     & 96 & 24 & 100 \\
            & DoA   &     & 32 & 2  & 1   \\
\bottomrule
\end{tabular*}
\end{table}
\subsection{Evaluation Metrics}\label{sec:eval_metrics} 

Across all scenarios, we evaluate attacks using three main metrics: 
(i) Attack Success Rate (ASR), which measures whether the targeted answers are successfully forced (or corrupted, in untargeted attacks); 
(ii) Collateral Damage (CDMG), which quantifies unintended changes to non-targeted answers within the same document; and 
(iii) ANLS-based scores, which capture token-level textual similarity and allow us to assess how predictions deviate from ground truth or move toward targeted answers.
We measure these metrics as a function of $B$, \ie, the number of QA pairs involved in the optimization objective.

\myparagraph{Attack Success Rate.} For all the attack scenarios, we measure the Attack Success Rate (ASR), defined as the percentage of cases where there is an exact match between the model answer and the one with which the adversarial example was optimized, \ie, $\hat{ \vct y} = \vct y^\star$ for targeted attacks and $\hat{ \vct y} \neq \vct y$ for the untargeted. 
As before, we use the index $i$ to indicate the sample, and $j$ to index the QA pairs related to that sample. 
The ASR for targeted attacks is then defined as:
\begin{equation}
\text{ASR} = \frac{1}{B} \sum_{i=1}^{N} \left( \prod_{j=1}^{B} \mathbf{1}\!\left[ \hat{\vct{y}}_{i,j} = \vct{y}^\star_{i,j} \right] \right)\label{eq:asr}
\end{equation}
where $B$ is the number of QA pairs targeted by the attack.
Note that, for multi-answer objectives, the success counts only if \textit{all} the QA pairs match the objective.
For the untargeted attacks, the ASR is computed using $\hat{\vct y}_{i, j} \neq \vct y_{i, j}$ in the indicator function.
Similarly to the other case, if multiple QA pairs are involved, the success is counted only if \textit{all} the answers are incorrect.

\myparagraph{Collateral Damage.} To evaluate how the attack affects the rest of the answers for a given document, \ie, the ones that are not in the optimization objective, we also compute the Collateral Damage (CDMG).
We define it as the percentage of QA pairs \textit{not} optimized whose prediction $\hat{\vct{y}}$ is wrong:
\begin{equation}
   \text{CDMG} = \frac{1}{C N} \sum_{i=1}^{N} \left( \sum_{j=1}^{C} \mathbf{1}\!\left[ \hat{\vct{y}}_{i,j} \neq \vct{y}^\star_{i,j} \right] \right)
   \label{eq:cdmg}
\end{equation}
where $C=M-B$ denotes the set of QA pairs not involved in the optimization of the attack.
As our dataset involves documents with $M=5$ QA pairs, when $B=5$, \ie, when we optimize all QA pairs for the document, this metric is undefined (and for this reason the curves are shown only for $C \in {1, \dots, 4}$). 
This metric is measured in the same way for targeted and untargeted attacks.

\myparagraph{ANLS.} For each attack scenario, we also measure the ANLS, with $\tau=0.5$. 
The ANLS is computed independently for each QA pair, and the final value is obtained by averaging across the full evaluation set. 
In our setting, we are interested in three distinct evaluations of this metric. 
The first is the average ANLS on the ground truth (ANLS-baseline), which measures the similarity between the model predictions and the original GT answers of the evaluation set (on all the $T$ QA pairs) . 
This is the starting point, \ie, the score without any perturbation.
The second, ANLS-B, measures the similarity between the predictions and the $B$ answers involved in the optimization.
The ANLS-B is expected to increase for targeted attacks, as the goal is to bring the answers closer to the targets.
Conversely, for untargeted attacks, we expect the metric to decrease (lower similarity).
Finally, we measure the ANLS-C, which measures the ANLS on the set of QA pairs kept out of the optimization. 
This metric should be correlated with the CDMG but captures if the attack is only disrupting  minimally (thus just enough to fail the exact match) or completely changing the other answers.


\myparagraph{Discussion.} For targeted, single-answer attacks, we expect high ASR and low CDMG, as the perturbation is optimized towards manipulating a specific QA pair.
On the other hand, for targeted multi-answer manipulation scenarios, high ASR might be more difficult to achieve, especially in the targeted attack (due to the fact that multiple QA pairs are optimized, and the ASR accounts for exact matches of \textit{all} answers).
CDMG may moderately increase because the visual features that influence one of the multiple answers optimized may overlap with those relevant to other questions, and therefore the optimization may inadvertently modify representations shared between multiple answers.
Finally, in the denial-of-answer setting, it is enough to disrupt the exact match of the answers with the ground truth (\ie, even by a single character). 
Moreover, in this case, the perturbation is explicitly designed to degrade the overall performance on the document rather than targeting specific answers.
Thus, the high CDMG is a \textit{desired} result.

\subsection{Experimental results}\label{sec:exp_results}
To evaluate the effectiveness of our method, for each model and each attack scenario, we show how the attack performance varies as the complexity of the attack objective increases, \ie, as a function of the number of QA pairs optimized. 

\begin{figure*}[t]
    \centering
    \includegraphics[width=1\linewidth]{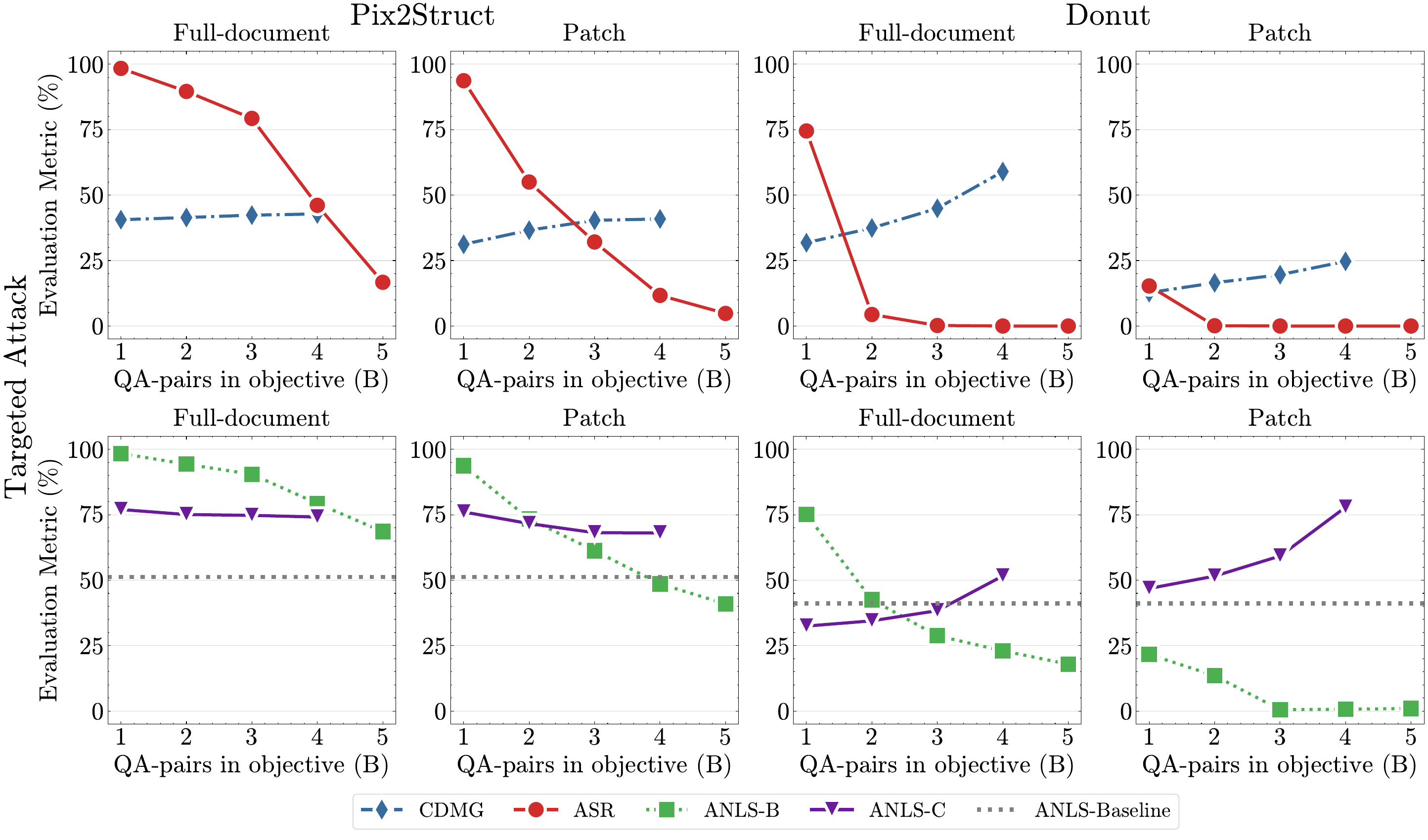}
    \caption{\textbf{Targeted attack results.} All metrics are reported for Pix2Struct and Donut across a different number of optimized QA pairs ($B$). We show ASR and the CDMG (top), and ANLS scores (bottom).}
    \label{fig:results_targeted}
\end{figure*}

\begin{figure*}[t]
    \centering
    \includegraphics[width=1\linewidth]{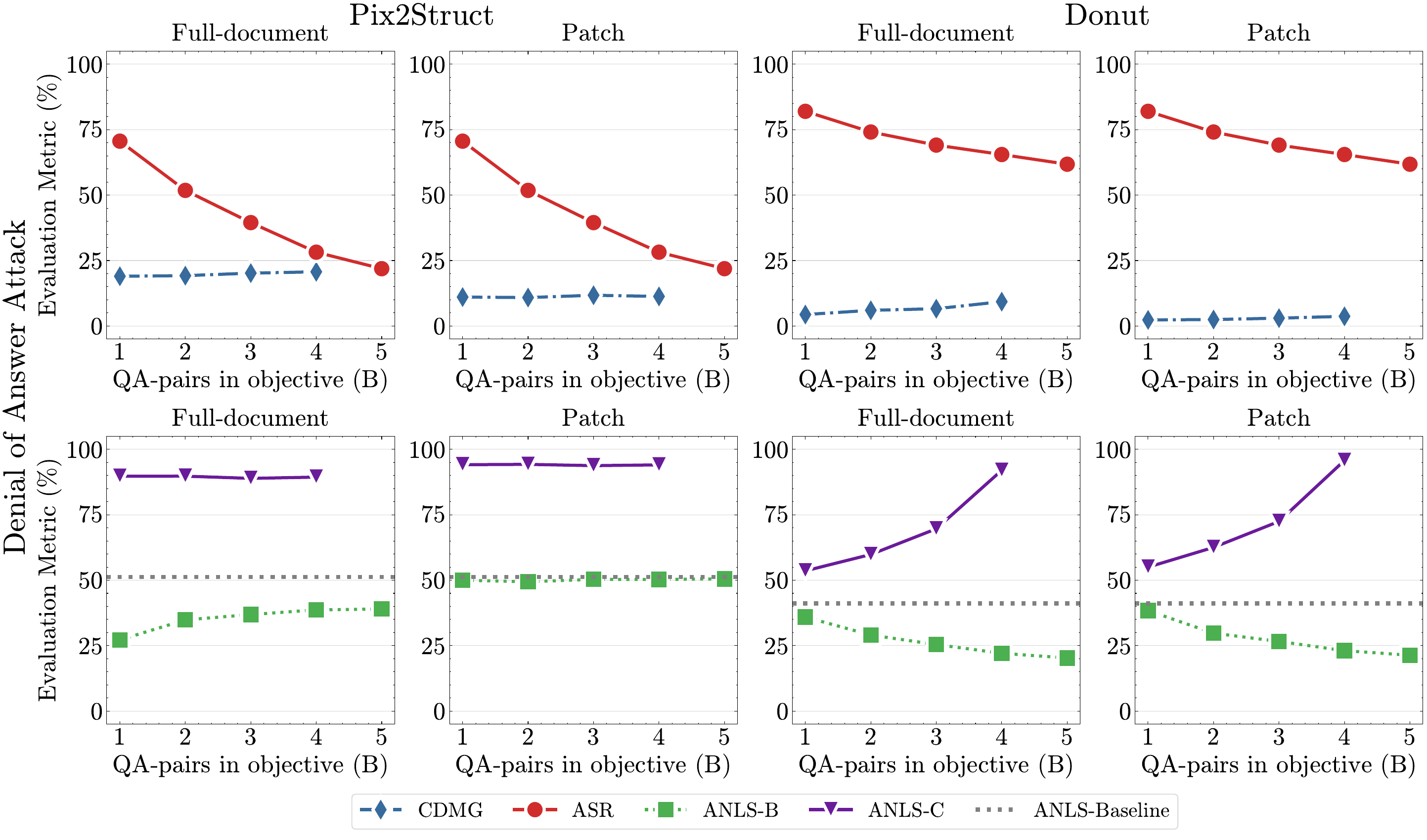}
    \caption{\textbf{Denial of Answer (DoA) results.} All metrics are reported for Pix2Struct and Donut across a different number of optimized QA pairs ($B$). We show ASR and the CDMG (top), and ANLS scores (bottom).}
    \label{fig:results_doa}
\end{figure*}


\myparagraph{Targeted Single-Answer manipulation ($B=1$).} 
Our first goal is to establish the effectiveness of the attack in the base scenario, \ie, the manipulation of a single answer. 
This case corresponds to the starting point of the curve in~\autoref{fig:results_targeted}, \ie, $B=1$. 
When perturbing the full document, our method achieves an ASR of nearly 100\% on Pix2Struct and just below 80\% on Donut.
The CDMG remains low for both models, i.e., the answers that are not affected by the attack tend to remain correct.
The same trends, with a smoother behavior, can be observed in the high ANLS-B~(\autoref{fig:results_doa}), confirming similarity of the answers to the targets, and high ANLS-C, meaning that the remaining answers remain close to the ground truth.
In the patch setting, the attack against Pix2Struct reaches almost 100\% ASR, while it remains below 20\% for Donut.
Again, the impact on the other answers is low for both models.
\textit{The attack effectively finds a perturbation that induces the model to output specific answers to a single question.}

\myparagraph{Targeted Multi-Answer manipulation ($B>1$).} 
We analyze how the attack behaves when the objective includes multiple QA pairs, \ie, $B>1$. 
Moving along the x-axis of~\autoref{fig:results_targeted}, the attack's effectiveness decreases with the number of QA pairs used in the optimization objective, reflecting a progressively more challenging scenario.
For Pix2Struct, the ASR remains high for small $B$ and gradually decreases as more QA pairs are included. 
The CDMG remains contained, indicating that the remaining QA pairs are barely impacted. 
The ANLS-B remains relatively high even as $B$ increases, indicating that the attack still improves similarity with the target answers across multiple QA pairs, but does not achieve exact match in many cases.
The ANLS-C remains high even with the addition of more questions in the optimization, confirming that the other answers are not affected.
For Donut, the ASR drops rapidly, approaching almost zero already at $B=2$, and the attack is largely ineffective for higher values of $B$.
The ANLS-B does not reach zero, suggesting that some generated answers still partially resemble the target sequences. 
In the patch attack scenario, a similar trend can be observed across all curves, with lower impact on the model due to the more restrictive scenario (which modifies only a portion rather the full document). 
For Pix2Struct, both the ASR and the ANLS-B are generally lower compared to the full-document setting. 
For Donut, the drop is even more pronounced, with ASR approaching $0$ already at $B=2$ and ANLS-B also decreasing.
\textit{Thus, jointly manipulating multiple answers substantially increases the attack difficulty.}
%

\myparagraph{Denial of Answer manipulation.} 
Next, we evaluate the Denial of Answer attack, \ie, the untargeted attack. 
The objective here is to maximize the loss with respect to the ground truth answer(s), forcing the model to provide unspecified incorrect outputs.
The results of this experiment are shown in \autoref{fig:results_doa}.
For the full-document attack, the method proves to be highly effective. 
The success rates in forcing a wrong answer (measured by ASR) is quite high with $B=1$, but decreases when the number of answers is higher because the attack fails to disrupt \textit{all} the QA pairs involved.
However, the drop in the efficacy of the attack for Donut is more contained than the one in the targeted case. 
This shows that untargeted disruption for this model is slightly easier.
The CDMG remains low for all conditions, meaning that the untargeted attack does not impact the remaining questions not directly optimized.
The ANLS metrics further confirm these results. 
The ANLS-B in this case measures how close the outputs are from the GT answers for which we want to elicit wrong outputs, and it becomes lower in the presence of the perturbation. 
The ANLS-C remains basically not impacted, except for Donut, in which they are affected for $B=1$, but less as the number of targets increases.
The patch attack scenario turns out to be similar to the full document setting, indicating that the additional constraint enforced by the patch is not limiting the effect of the attack.
\textit{Denial-of-Answer attacks corrupt the targeted answers with minimal impact to the rest of the document.}


\section{Related Work}
\label{sec:related}
We summarize related research, structured into attacks that aim to adversarially disrupt systems by attacking the preprocessing part (either the OCR in OCR-based systems or the downsampling of image data in general), adversarial attacks against the ML models in isolation, and other attacks against multimodal models.

\myparagraph{Adversarial attacks against the preprocessing.}
Several works have studied how non-differentiable or lossy preprocessing steps can be exploited to attack vision models. 
Song~\etal\cite{song2018fooling} demonstrate that Optical Character Recognition (OCR) systems are vulnerable to adversarial perturbations that survive text extraction. 
Beyond OCR, preprocessing steps such as resizing or compression can themselves be targeted. 
Xiao et al.~\cite{xiao2019seeing} introduce image-scaling attacks, where perturbations become adversarial after resizing. 
Quiring et al.~\cite{quiring2020adversarial} provide a broader taxonomy of weaknesses arising from resampling and quantization in vision pipelines. 
In contrast, we study end-to-end document understanding models, where perturbations must propagate through integrated visual–textual reasoning rather than isolated preprocessing steps.

\myparagraph{Attacks against multimodal models.} 
Bailey \etal~\cite{bailey2024image} show that adversarial images can hijack the output of generative multimodal models at inference time, with the objective of steering high-level generative behavior.
Similarly, Cui et al.~\cite{cui2024robustness} and Schlarmann \etal~\cite{schlarmann2023adversarial} conduct a systematic robustness study, confirming that LMMs are broadly vulnerable to adversarial images. 
In the domain of documents, recent works explored multimodal inference risks: DocMIA~\cite{nguyen2025docmia} highlights that sensitive document content can be leaked through model outputs.
However, these works primarily study broad alignment bypasses or information leakage.
Moreover, they don't target multiple answers at the same time.
In contrast, our work focuses on crafting end-to-end differentiable attacks that manipulate multiple fine-grained answers in the document analysis pipeline, highlighting a previously unexplored vulnerability surface for this domain.

\myparagraph{Jailbreaks in autoregressive models.} 
Another line of work targets alignment mechanisms in LLMs and LMMs. 
Greedy Coordinate Gradient (GCG)~\cite{zou2023universal} performs gradient-based optimization over discrete tokens to induce jailbreaks, and Carlini et al.~\cite{carlini2023aligned} show that visual adversarial inputs can bypass multimodal safety training. 
In contrast to jailbreak attacks on LLMs, that aim to bypass content filters and alignment mechanisms, our work focuses on controlling or disrupting specific downstream answers. Importantly, the resulting outputs are plausible and policy-compliant (aligned) in form.

\section{Conclusions}
\label{sec:conclusion}

In this work, we have outlined the threat model for multimodal DocVQA systems, and demonstrated that they are vulnerable to end-to-end adversarial perturbations that can be crafted to produce incorrect or targeted answers. 
Our experiments show that small, visually imperceptible (full-document) or still inconspicuous and localized (patch) perturbations are often sufficient to substantially degrade model performance or achieve target responses, and that these attacks can disrupt multiple question–answer pairs on the same document simultaneously. 
These results highlight a previously underexplored attack surface on this domain.

We suggest several directions for future research. 
First, our results assume a fully white-box threat model; relaxing this assumption by averaging over variations of the same question or by optimizing against ensembles/surrogates of models would make the threat model more realistic. 
Second, many real-world document tasks operate on multi-page documents; extending attacks to multi-page DocVQA remains an important open problem.  
Third, developing \emph{universal} perturbations that generalize across documents (for example, via patch-based methods that resemble watermarks) is a promising but technically nontrivial direction. 
Finally, future work should investigate practical defenses and robust evaluation protocols: realistic preprocessing hardening, model-side detection of anomalous inputs, adversarial training adapted to document modalities. 
We hope this study draws attention to the concrete risks of adversarial manipulation in DocVQA systems and stimulates development of principled, deployable mitigations.


\section*{Acknowledgments}
This work has been partly supported by the EU-funded Horizon Europe projects ELSA (GA no.
101070617); and by
the projects SERICS (PE00000014) and FAIR (PE00000013) under the MUR National Recovery
and Resilience Plan funded by the European Union - NextGenerationEU; and by project PID2023-146426NB-100 funded by MCIU/AEI/10.13039/501100011033 and FEDER, UE.

\bibliographystyle{IEEEtran}
\bibliography{bibliography}

@String(ICCV= {Int. Conf. Comput. Vis.})

@String(ECCV= {Eur. Conf. Comput. Vis.})

@String(ICLR = {Int. Conf. Learn. Represent.})

@String(ICCV  = {ICCV})

@String(ECCV  = {ECCV})

@String(ICLR  = {ICLR})

@inproceedings{kim2022donut,
  title     = {OCR-Free Document Understanding Transformer},
  author    = {Kim, Geewook and Hong, Teakgyu and Yim, Moonbin and Nam, JeongYeon and Park, Jinyoung and Yim, Jinyeong and Hwang, Wonseok and Yun, Sangdoo and Han, Dongyoon and Park, Seunghyun},
  booktitle = ECCV,
  year      = {2022}
}

@inproceedings{lee2023pix2struct,
author = {Lee, Kenton and Joshi, Mandar and Turc, Iulia and Hu, Hexiang and Liu, Fangyu and Eisenschlos, Julian and Khandelwal, Urvashi and Shaw, Peter and Chang, Ming-Wei and Toutanova, Kristina},
title = {Pix2Struct: screenshot parsing as pretraining for visual language understanding},
year = {2023},
publisher = {JMLR.org},
booktitle = ICML,
articleno = {780},
}

@inproceedings{biggio2013evasion,
  title={Evasion attacks against machine learning at test time},
  author={Biggio, Battista and Corona, Igino and Maiorca, Davide and Nelson, Blaine and {\v{S}}rndi{\'c}, Nedim and Laskov, Pavel and Giacinto, Giorgio and Roli, Fabio},
  booktitle={Machine learning and knowledge discovery in databases: European conference, ECML PKDD},
  year={2013},
}

@inproceedings{szegedy2014intriguing,
  title={Intriguing properties of neural networks},
  author={Szegedy, Christian and Zaremba, Wojciech and Sutskever, Ilya and Bruna, Joan and Erhan, Dumitru and Goodfellow, Ian and Fergus, Rob},
  booktitle=ICLR,
  year={2014}
}

@inproceedings{tito2024privacy,
  title = {Privacy-Aware Document Visual Question Answering},
  author = {Tito, Rubén and Nguyen, Khanh and Tobaben, Marlon and Kerkouche, Raouf and Souibgui, Mohamed Ali and Jung, Kangsoo and J\"alk\"o, Joonas and Poulain D'Andecy, Vincent and Joseph, Aurélie and Kang, Lei and Valveny, Ernest and Honkela, Antti and Fritz, Mario and Karatzas, Dimosthenis},
  booktitle = {Proc. of the International Conference on Document Analysis and Recognition (ICDAR)},
  year = {2024}
}

@inproceedings{mathew2021docvqa,
  author    = {Mathew, M. and Karatzas, D. and Jawahar, C.},
  title     = {DocVQA: A Dataset for VQA on Document Images},
  booktitle = {Proc. of the IEEE/CVF Winter Conference on Applications of Computer Vision},
  year      = {2021}
}

@inproceedings{biten2019scene,
  title={Scene text visual question answering},
  author={Biten, Ali Furkan and Tito, Ruben and Mafla, Andres and Gomez, Lluis and Rusinol, Mar{\c{c}}al and Valveny, Ernest and Jawahar, CV and Karatzas, Dimosthenis},
  booktitle={Proc. of the IEEE/CVF international conference on computer vision},
  year={2019}
}

@article{brown2017adversarial,
  title={Adversarial patch},
  author={Brown, Tom B and Man{\'e}, Dandelion and Roy, Aurko and Abadi, Mart{\'\i}n and Gilmer, Justin},
  journal={arXiv preprint arXiv:1712.09665},
  year={2017}
}

@inproceedings{cui2024robustness,
  title={On the robustness of large multimodal models against image adversarial attacks},
  author={Cui, Xuanming and Aparcedo, Alejandro and Jang, Young Kyun and Lim, Ser-Nam},
  booktitle={Proc. of the IEEE/CVF Conference on Computer Vision and Pattern Recognition},
  year={2024}
}

@article{carlini2023aligned,
  title={Are aligned neural networks adversarially aligned?},
  author={Carlini, Nicholas and Nasr, Milad and Choquette-Choo, Christopher A and Jagielski, Matthew and Gao, Irena and Koh, Pang Wei W and Ippolito, Daphne and Tramer, Florian and Schmidt, Ludwig},
  journal={Advances in Neural Information Processing Systems},
  year={2023}
}

@inproceedings{quiring2020adversarial,
author = {Erwin Quiring and David Klein and Daniel Arp and Martin Johns and Konrad Rieck},
title = {Adversarial Preprocessing: Understanding and Preventing {Image-Scaling} Attacks in Machine Learning},
booktitle = {29th USENIX Security Symposium},
year = {2020},
isbn = {978-1-939133-17-5},
}

@inproceedings{nguyen2025docmia,
  title={DocMIA: Document-Level Membership Inference Attacks against DocVQA Models},
  year = {2025},
  author={Nguyen, Khanh and Kerkouche, Raouf and Fritz, Mario and Karatzas, Dimosthenis},
  booktitle=ICLR
}

@inbook{dong2025position,
publisher={International Conference on Document Analysis and Recognition},
author = {Dong, Qi and Kang, Lei and Pintor, Maura and Karatzas, Dimosthenis},
year = {2025},
month = {09},
title = {Position-Aware Stamp-Like Adversarial Attack for Document Classification},
isbn = {978-3-032-04626-0},
doi = {10.1007/978-3-032-04627-7_17},
}

@inproceedings{xiao2019seeing,
  title={Seeing is not believing: Camouflage attacks on image scaling algorithms},
  author={Xiao, Qixue and Chen, Yufei and Shen, Chao and Chen, Yu and Li, Kang},
  booktitle={28th USENIX Security Symposium},
  year={2019}
}

@inproceedings{madry2018towards,
  title={Towards Deep Learning Models Resistant to Adversarial Attacks},
  author={Madry, Aleksander and Makelov, Aleksandar and Schmidt, Ludwig and Tsipras, Dimitris and Vladu, Adrian},
  booktitle=ICLR,
  year={2018}
}

@inproceedings{dong2018boosting,
  title={Boosting adversarial attacks with momentum},
  author={Dong, Yinpeng and Liao, Fangzhou and Pang, Tianyu and Su, Hang and Zhu, Jun and Hu, Xiaolin and Li, Jianguo},
  booktitle=ICCV,
  year={2018}
}

@article{zou2023universal,
  title={Universal and transferable adversarial attacks on aligned language models},
  author={Zou, Andy and Wang, Zifan and Carlini, Nicholas and Nasr, Milad and Kolter, J Zico and Fredrikson, Matt},
  journal={arXiv preprint arXiv:2307.15043},
  year={2023}
}

@article{song2018fooling,
  title={Fooling OCR systems with adversarial text images},
  author={Song, Congzheng and Shmatikov, Vitaly},
  journal={arXiv:1802.05385},
  year={2018}
}

@InProceedings{bailey2024image,
  title = 	 {Image Hijacks: Adversarial Images can Control Generative Models at Runtime},
  author =       {Bailey, Luke and Ong, Euan and Russell, Stuart and Emmons, Scott},
  booktitle = 	 ICML,
  year = 	 {2024},
  editor = 	 {Salakhutdinov, Ruslan and Kolter, Zico and Heller, Katherine and Weller, Adrian and Oliver, Nuria and Scarlett, Jonathan and Berkenkamp, Felix},
  volume = 	 {235},
}

@inproceedings{schlarmann2023adversarial,
  title={On the Adversarial Robustness of Multi-Modal Foundation Models},
  author={Schlarmann, Christian and Hein, Matthias},
  booktitle={2023 IEEE/CVF International Conference on Computer Vision Workshops (ICCVW)},
  year={2023},
  organization={IEEE}
}

\end{document}